\documentclass[runningheads]{llncs}
\usepackage[T1]{fontenc}
\usepackage{graphicx}
\usepackage{booktabs}
\usepackage{amsmath}
\usepackage{multirow}
\usepackage{array}
\usepackage{bbding}
\usepackage{url}

\begin{document}

\title{Analyzing Atypical Local Features in Museum Artifacts using Generative Image Inpainting}
\titlerunning{Analyzing Atypical Local Features in Museum Artifacts}
\author{
Maho~Kinoshita\inst{1}\Envelope\orcidID{0009-0007-9914-7609} \and \\
Wakana~Kuwata\inst{1}\orcidID{0009-0009-4612-4467} \and \\
Ryota~Mibayashi\inst{2}\orcidID{0009-0005-8995-5579} \and \\
Hiroaki~Ohshima\inst{1}\orcidID{0000-0002-9492-2246}}
\authorrunning{M. Kinoshita et al.}

\institute{University of Hyogo, Hyogo, Japan\\
Kobe University, Hyogo, Japan\\}
\authorrunning{M. Kinoshita et al.}

\maketitle

\begin{abstract}

This study proposes a method for analyzing local visual features in museum artifacts that may appear atypical or unexpected.
In museum artifacts, objects in the same category often share a similar overall shape, 
while local details such as patterns, decorations, and facial expressions may reflect regional or cultural characteristics.
To find such features, it is necessary to define a reference for what appearance is generally expected within the category.
In this study, we use the appearance distribution learned 
by an image generation model to represent what is generally expected for an artifact category.
Specifically, for each target region, we exclude images from that region from the training data 
and fine-tune a generative model using only images from the remaining regions.
This produces a region excluded model that has not directly learned the visual features specific to the target region.
We then mask selected parts of artifact images from the target region 
and ask the region-excluded model to complete the missing regions.
If the completed result differs greatly from the original image, 
the masked part may contain features that are difficult to predict from the general appearance learned from other regions.
We measure this difference using LPIPS and treat it as reconstruction difficulty for the generative model.
We conduct experiments on images of \textit{kokeshi dolls}, \textit{drums}, 
and \textit{masks} from the collection of the National Museum of Ethnology.
For each category, we select several target regions, 
build region-excluded LoRA models, and perform inpainting on predefined local parts.
The results show that typical shapes and simple decorations are often reconstructed consistently, 
whereas some region-specific colors, patterns, cord arrangements, and painted details are replaced with more common appearances.
These results suggest that region-excluded inpainting can serve as an exploratory tool 
for finding region-specific local visual features in museum artifact images.

\end{abstract}
\section{Introduction}

When people find a feature that differs from what they expect in a familiar object,
they often pay attention to that part.
Such features can leave a strong impression
and may become cues for memory and learning.
In psychology and cognitive science,
features that violate predictions or expectations are regarded as important factors
related to attention, learning, and memory~\cite{Itti2005Bayesian}.

Objects can be unexpected in several ways,
such as in their function, region, period, or material.
Among these aspects, this study focuses on visual unexpectedness,
which people can often notice quickly.
Visual unexpectedness does not only occur when the overall shape or structure
of an object is very different from known examples.
It can also appear when the overall shape is typical,
but local details such as patterns, decorations, or facial expressions
differ from what is expected.

\begin{figure}[tb]
    \centering
    \includegraphics[scale = 0.24]{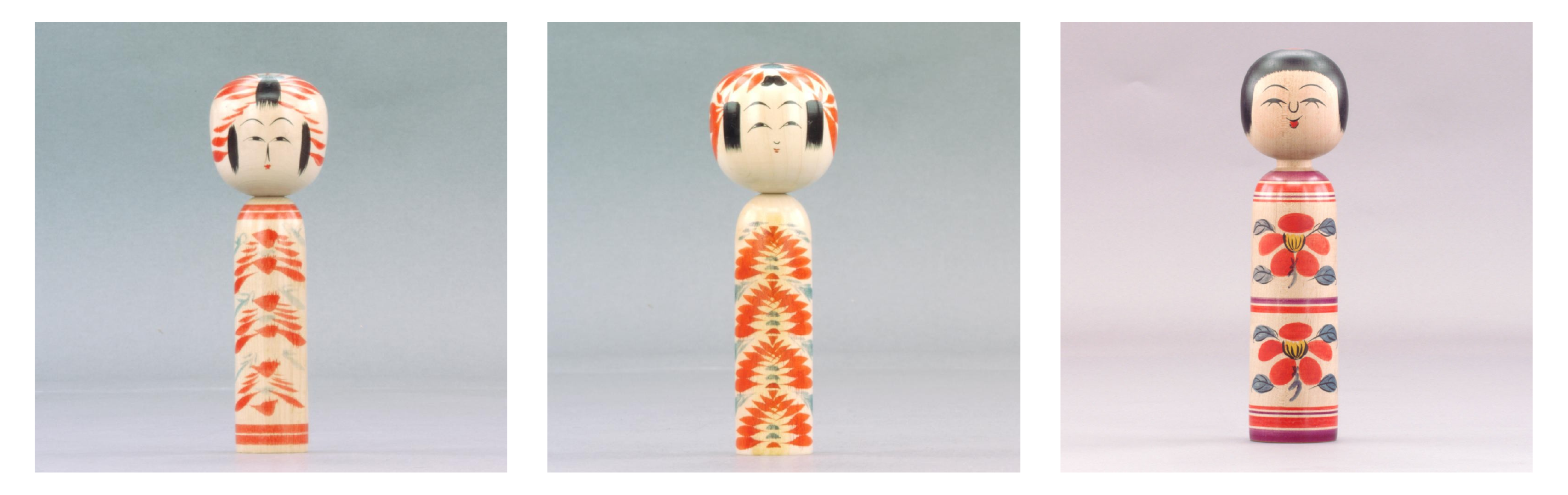}
    \caption{Examples of \textit{kokeshi dolls} from different regions.
            Although they share a similar overall structure, local features such as hair,
            neck stripes, and body decorations vary across regions.}
    \label{fig:kokeshis}
\end{figure}
This study focuses on the latter type of visual unexpectedness:
local and atypical visual features embedded in an otherwise typical overall shape.
We use images of \textit{kokeshi dolls}, \textit{drums}, and \textit{masks}
from the collection of the National Museum of Ethnology.
\textit{Kokeshi dolls} are traditional Japanese wooden dolls,
whereas \textit{drums} and \textit{masks} are collected across diverse regions and cultures.
Artifacts in the same category often share similar overall structures,
while their details, such as facial expressions, body patterns,
string arrangements, and ornaments, differ across regions.
These artifacts are therefore suitable for analyzing region-specific local features
that appear within typical object shapes.
For example, consider the \textit{kokeshi dolls} from several regions shown in Fig~\ref{fig:kokeshis}.
The hair, neck stripes, and floral patterns on the body vary depending on the region.
To quantify local visual unexpectedness,
we first need a reference for what appearance is expected.
Unexpectedness is not a fixed property of an object itself.
Rather, it arises from a gap between the object and expectations
formed within a certain category or context.
Therefore, to examine which parts of an artifact image are visually unexpected,
we need to model what appearance is generally or typically expected
for that category.

Recent image generation models learn visual regularities from large image datasets
and can generate plausible appearances for a given context.
This property suggests that such models can be used
as computational approximations of visual expectations.
The key idea of this study is to use a region excluded generative model
as a visual expectation model.
For each target region, we exclude images from that region
and fine-tune the model using only artifact images from other regions.
This creates a model that has not directly learned
the region-specific features of the target region.
If this model cannot naturally inpaint a part of a target region artifact,
the part may contain local visual features
that deviate from the general appearance distribution learned by the model.

\begin{figure}[tb]
  \centering
  \includegraphics[width=0.8\textwidth]{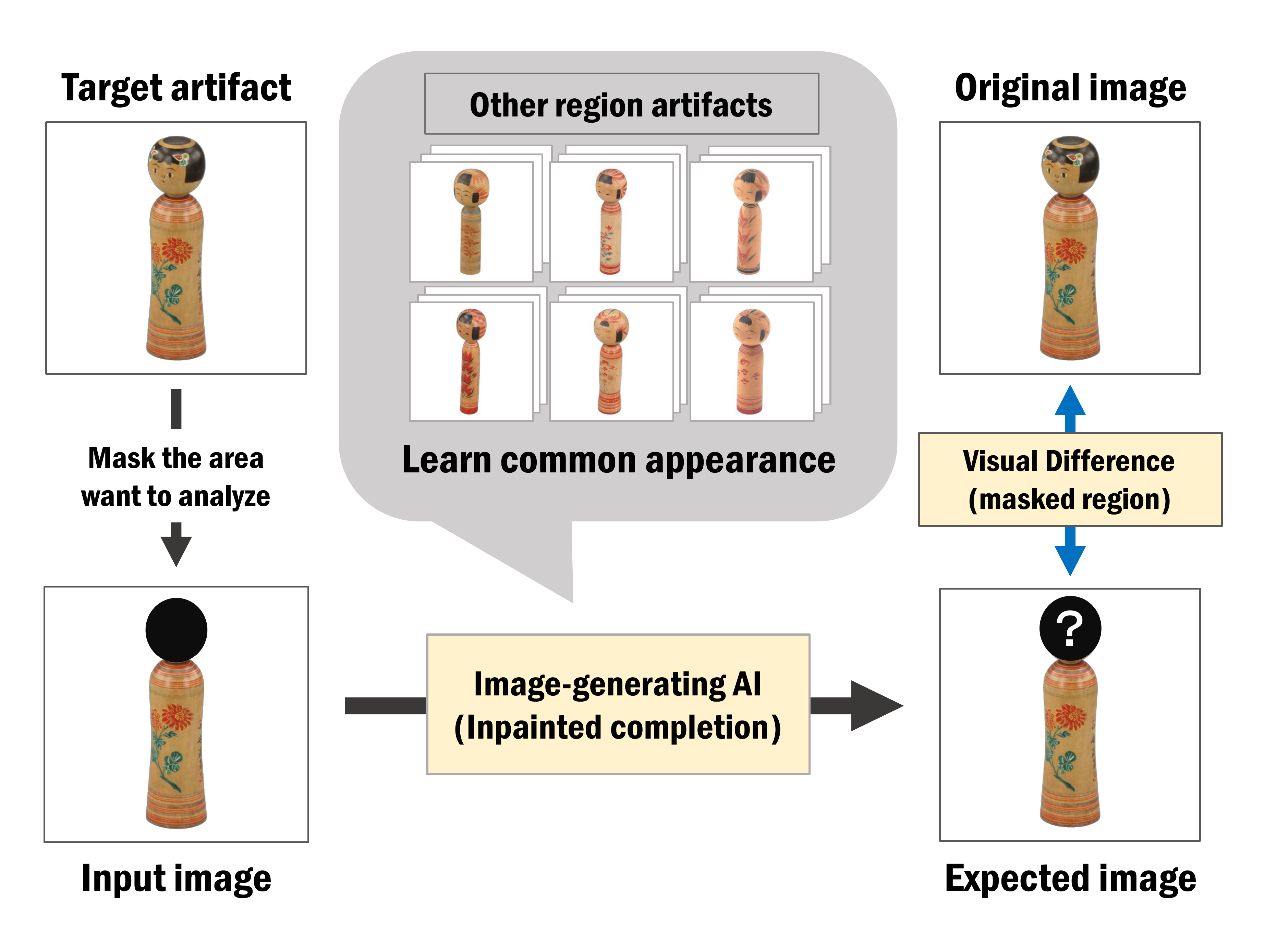}
  \caption{Conceptual overview of the proposed method.}
  \label{fig:concept}
\end{figure}

Fig.~\ref{fig:concept} shows the conceptual overview of the proposed method.
Using the region excluded model,
we inpaint masked parts of artifact images from the target region
and measure the difference between the original and inpainted regions.
We treat this difference as reconstruction difficulty
relative to the model's learned appearance distribution,
and use it as a cue for local visual unexpectedness.

The proposed method consists of the following three steps:
\begin{enumerate}
  \item We fine-tune a generative model using artifact images
        from regions other than the target region.
  \item We mask local parts of target region artifact images
        and inpaint them using the region excluded model.
  \item We compare the original and inpainted regions
        and measure their perceptual difference using LPIPS.
\end{enumerate}

We define this LPIPS based difference as reconstruction difficulty
for the generative model.
A higher score indicates that the local region is more difficult to predict
from the general appearance distribution learned without the target region.
We analyze whether this score can capture region-specific local visual features
by examining both quantitative LPIPS scores and qualitative inpainting results.
The main contributions of this study are as follows.
\begin{itemize}
  \item \textbf{Inpainting based estimation of local visual unexpectedness.}
        We propose a framework for measuring local visual unexpectedness
        based on image inpainting by a generative model.
    \item \textbf{Region excluded expectation modeling.}
        We apply LoRA fine-tuning with the target region excluded 
        and analyze region specific visual features
        as deviations from learned visual expectations.
    \item \textbf{Evaluation on multiple museum artifact categories.}
        Through experiments on \textit{kokeshi dolls}, \textit{drums}, and \textit{masks},
        we show that the proposed score highlights local patterns, decorations,
        and structural details that are difficult to reconstruct
        from region excluded appearance distributions.
\end{itemize}

\section{Related Work}

This study estimates local visual unexpectedness in museum artifacts
as reconstruction difficulty during image inpainting by a generative model.
This section reviews expectation based perception and unexpectedness,
AI applications for cultural heritage materials,
inpainting and reconstruction based deviation analysis,
and perceptual metrics for generated images.

\subsection{Expectation-Based Perception, Surprise, and Interest}

Unexpectedness has been studied as a gap between predictions or expectations
and actual perceptual input.
Humans predict the outside world based on past experience and context,
and stimuli that violate these predictions tend to attract attention.
This view is closely related to predictive processing and predictive coding
~\cite{Friston2004cortical,Friston2009free,Rao1999predictive}.

Computational studies have modeled surprise through Bayesian surprise,
which measures belief updates caused by observations,
and through information theoretic views of prediction difficulty
~\cite{ITTI2009Bayesian,Shannon1948math}.
In visual attention research, saliency models have also been proposed
to estimate attention attracting regions from low level visual features
~\cite{Itti1998Saliency,Borji2013Saliency}.

Visual unexpectedness is also related to curiosity, interest, and learning.
Studies on curiosity and interest have shown that knowledge gaps,
novelty, complexity, and comprehensibility affect exploration and memory
~\cite{Loewenstein1994curiosity,Kang2009curiosity,Gruber2014curiosity,Silvia2005interest,Silvia2008interest}.
In this study, reconstruction difficulty is not treated as curiosity or interest itself,
but as a computational cue for expectation violation
that may lead to curiosity or interest.

\subsection{AI for Cultural Heritage and Museum Artifact Analysis}

In recent years, AI has been increasingly used
for cultural heritage and museum materials.
Existing studies have addressed image classification, material retrieval,
metadata annotation, visualization of cultural meaning,
support for preservation and restoration,
and museum experience or learning support
~\cite{Fiorucci2020heritage,Goodarzi2023designing,Quan2024batikculture,Falk2005learning,Hein1998Museum}.

Cultural heritage materials contain diverse visual features
related to region, period, function, material, and decoration,
making them important targets for machine learning based analysis.
However, most existing studies focus on classification, identification,
semantic annotation, or preservation support.
Relatively few studies quantify how unexpected local visual features
of artifacts may appear to viewers.

This study addresses this gap by analyzing local visual features
of museum artifacts as deviations from expectations
formed from region and category.

\subsection{Inpainting and Reconstruction-Based Deviation Analysis}

Diffusion models have been widely used for high quality image generation.
Based on DDPMs and Latent Diffusion Models,
they have been applied to high resolution image generation and image editing
~\cite{Ho2020DDPM,Rombach2022sd}.

Image inpainting masks part of an input image
and fills the missing part to match the surrounding context.
Earlier methods used GANs or convolutional networks,
while recent diffusion based methods such as RePaint and Palette
have enabled more flexible and high quality completion
~\cite{Pathak2016context,Yu2018contextual,Liu2018partialconv,Lugmayr2022RePaint,Saharia2022Palette}.

In this study, inpainting is not used only as image repair.
Instead, it is used to observe what appearance
the generative model expects for the masked part.
A large difference between the original and inpainted part
indicates that the part is difficult for the model to reconstruct
and may deviate from expectations based on the learned visual distribution.

This idea is methodologically related to reconstruction based anomaly detection,
which assumes that a model trained on normal data
cannot accurately reconstruct patterns outside the learned distribution
~\cite{schlegl2017anogan,Wyatt2022AnoDDPM,Ancha2025AD,jing2025DMADsurvey}.
However, this study does not aim to detect defects or abnormalities.
Instead, it analyzes how much region-specific visual features
in cultural artifacts deviate from learned expectations.

\subsection{Perceptual Metrics for Reconstruction Difficulty}

When measuring differences between generated and original images,
metrics such as MAE, PSNR, and SSIM have been used
based on pixel values or structural similarity
~\cite{wang2004SSIM}.
LPIPS measures perceptual similarity using distances
in a deep feature space and is often used as a distance measure
closer to human perceptual judgments
~\cite{zhang2018LPIPS}.

In this study, LPIPS is not used as a direct definition of human unexpectedness.
Instead, it is used as a computational measure of reconstruction difficulty
for the generative model.
We then use this reconstruction difficulty to analyze local visual features
that are not easily reproduced by the region-excluded generative model.

\section{Problem Definition}

This study aims to estimate the unexpectedness of local visual features in artifact images.
We view unexpectedness not as an attribute of the object itself,
but as a gap from the visual expectations formed for the category.
Therefore, this study uses the visual distribution learned by an image generation model
as a representation of expectations.
We evaluate local visual unexpectedness from the difference between the inpainting result and the original image.
Let $C$ be the target category,
and let $R=\{r_1,\ldots,r_m\}$ be the set of regions used for analysis.
For a region $r_i \in R$, we denote by $M_{-r_i}$ a generative model
built by excluding images from region $r_i$ from the training data.
This model is regarded as reflecting the visual expectations of an observer
who has not seen the features of region $r_i$.
The input is an artifact image $x$ from region $r_i$ and a mask $m$
that indicates the part to be analyzed.
Using the generative model, we inpaint the masked part and obtain a generated image $x_{\mathrm{gen}}$.
We then measure the perceptual difference between the original image and the generated image using LPIPS~\cite{zhang2018LPIPS}.
The visual unexpectedness score of part $p$ in image $x$ from part $r_i$
is defined as the average LPIPS distance over multiple inpainting results:

\begin{equation}
S(x, p, r_i) = \frac{1}{N} \sum_{n=1}^{N}
\mathrm{LPIPS}\left(
\mathrm{crop}(x, p),
\mathrm{crop}(x_{\mathrm{gen}}^{n}, p)
\right)
\label{eq:local_score}
\end{equation}
Here, $x_{\mathrm{gen}}^{n}$ denotes the $n$th inpainting result,
and $\mathrm{crop}(x,p)$ denotes the cropped image part corresponding to part $p$.
In this study, we set $N=5$.
A higher visual unexpectedness score means that the model had more difficulty reconstructing the local part.
In other words, the part is likely to contain features that are hard to predict
from the general visual distribution learned by the model.
We also define the visual unexpectedness score of the whole artifact
as the average over all parts:

\begin{equation}
S(x, r_i) = \frac{1}{|P_C|}
\sum_{p \in P_C} S(x, p, r_i)
\label{eq:global_score}
\end{equation}
This study examines how well the calculated visual unexpectedness score captures
region-specific visual features.
We also analyze what kinds of local visual features 
tend to produce high reconstruction difficulty scores.

\section{Museum Artifact Dataset and Region Selection}
\subsection{Dataset Construction}

The data used in this study were built from specimen image data owned by the National Museum of Ethnology
\footnote{\url{https://www.minpaku.ac.jp/}} (Minpaku).
The original data contain many types of ethnological materials.
From this data, we collected artifacts and their corresponding images by keyword matching.
As a result, we obtained 593 images for \textit{kokeshi dolls}, 929 images for \textit{drums},
and 4,629 images for \textit{masks}.
Each image has metadata about its production area.
The metadata vary in detail: some are given at the prefecture level,
some at a broader regional level, and some are unknown.
Using this information, the dataset has a structure that allows us to analyze
regional style differences to some extent for each artifact type.
\textit{Kokeshi dolls} are artifacts with strong regional lineages.
However, some images include designs that differ from such traditional styles,
such as bodies with extremely rounded shapes or heads with very distinctive hairstyles.
These examples are not suitable for extracting regional features.
Drums and masks also include artifacts that differ greatly from traditional styles.

\begin{figure}[tb]
  \centering
  \includegraphics[width=0.8\textwidth]{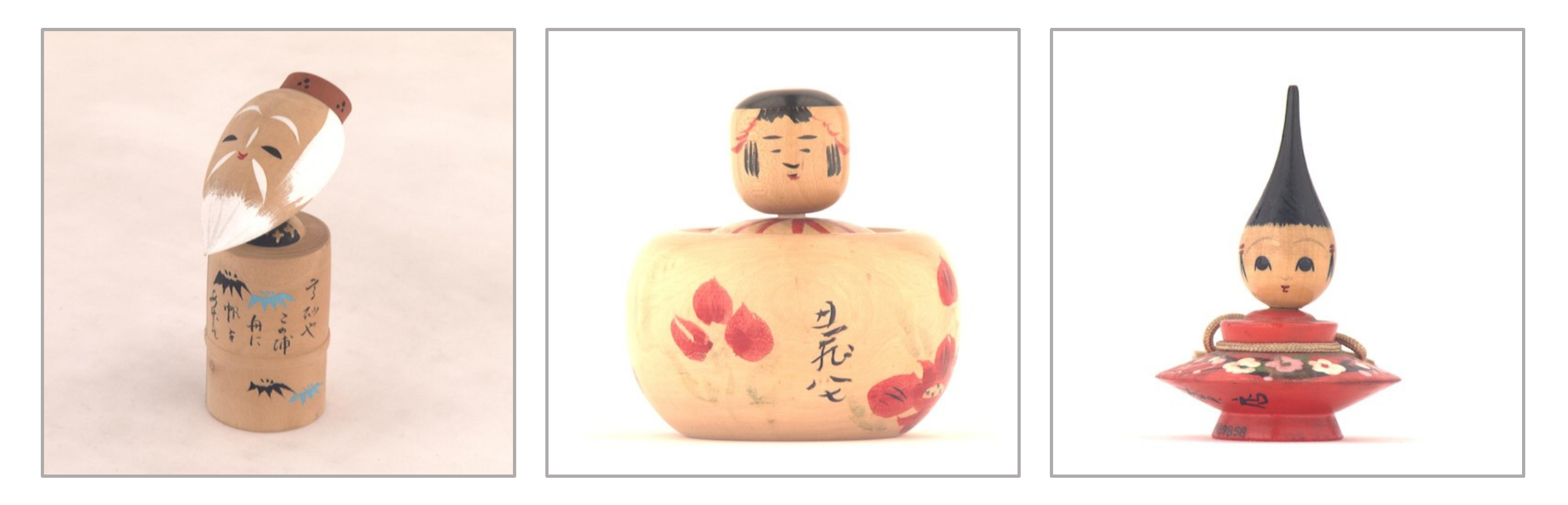}
  \caption{Examples of removed images}
  \label{fig:no_kokeshis}
\end{figure}

To automatically remove artifacts that strongly differ from the basic shape,
as well as unrelated noisy images,
we used VisualRank based on the silhouette of each artifact to measure shape typicality.
We calculated a typicality score for all samples and removed images with low scores.
For the drum category, we removed the bottom 40\% of images by score.
Examples of removed images are shown in Fig.~\ref{fig:no_kokeshis}.
After this refinement, the final dataset contained 536 kokeshi images,
624 drum images, and 2,479 mask images.

To reduce the effect of differences in background and scale on unexpectedness estimation,
we preprocessed all images.
Specifically, we removed the background and normalized the image size
so that each artifact was placed at the center of a same sized canvas.

\subsection{Region Selection for Evaluation}

To cover regional style differences in each category,
we selected four regions with different visual styles for each category.
For \textit{kokeshi dolls}, we selected four prefectures in the Tohoku region of Japan:
Yamagata, Fukushima, Miyagi, and Aomori.
For \textit{drums}, we selected four countries: Indonesia, Nepal, Iran, and Morocco.
For \textit{masks}, we selected four countries: Bhutan, Vietnam, Kenya, and Mexico.
These regions have enough images to select representative samples for evaluation.
They also show differences in appearance, such as decorations, shapes, and patterns.
Therefore, we judged them to be suitable for comparing region-specific visual features.

From each region, we selected five representative images with the highest typicality scores
using VisualRank.
This makes it possible to focus the analysis on typical examples from each region.
The evaluation dataset consists of 60 images in total:
five images from each of four regions in each of the three categories.
Table~\ref{tab:dataset} shows the selected regions and the training data size used for each LoRA model.

\begin{table}[tb]
    \caption{Selected regions and image counts used in the region excluded setting.}
    \label{tab:dataset}
    \centering
        \begin{tabular}{llr}
        \hline \hline
        Category & Region & Number of images \\
        \hline
        Kokeshi & Yamagata  & 32 \\
                & Fukushima & 22 \\
                & Miyagi    & 98 \\
                & Aomori    & 13 \\
                & Total for other regions & 191 \\
        \hline
        Drum    & Indonesia & 33 \\
                & Nepal     & 6 \\
                & Iran      & 7 \\
                & Morocco   & 13 \\
                & Total for other regions  & 870 \\
        \hline
        Mask    & Bhutan    & 31 \\
                & Vietnam   & 16 \\
                & Kenya     & 10 \\
                & Mexico    & 447 \\
                & Total for other regions  & 2,273 \\
        \hline
        \end{tabular}
\end{table}

\begin{figure}[tb]
  \centering
  \includegraphics[width=0.8\textwidth]{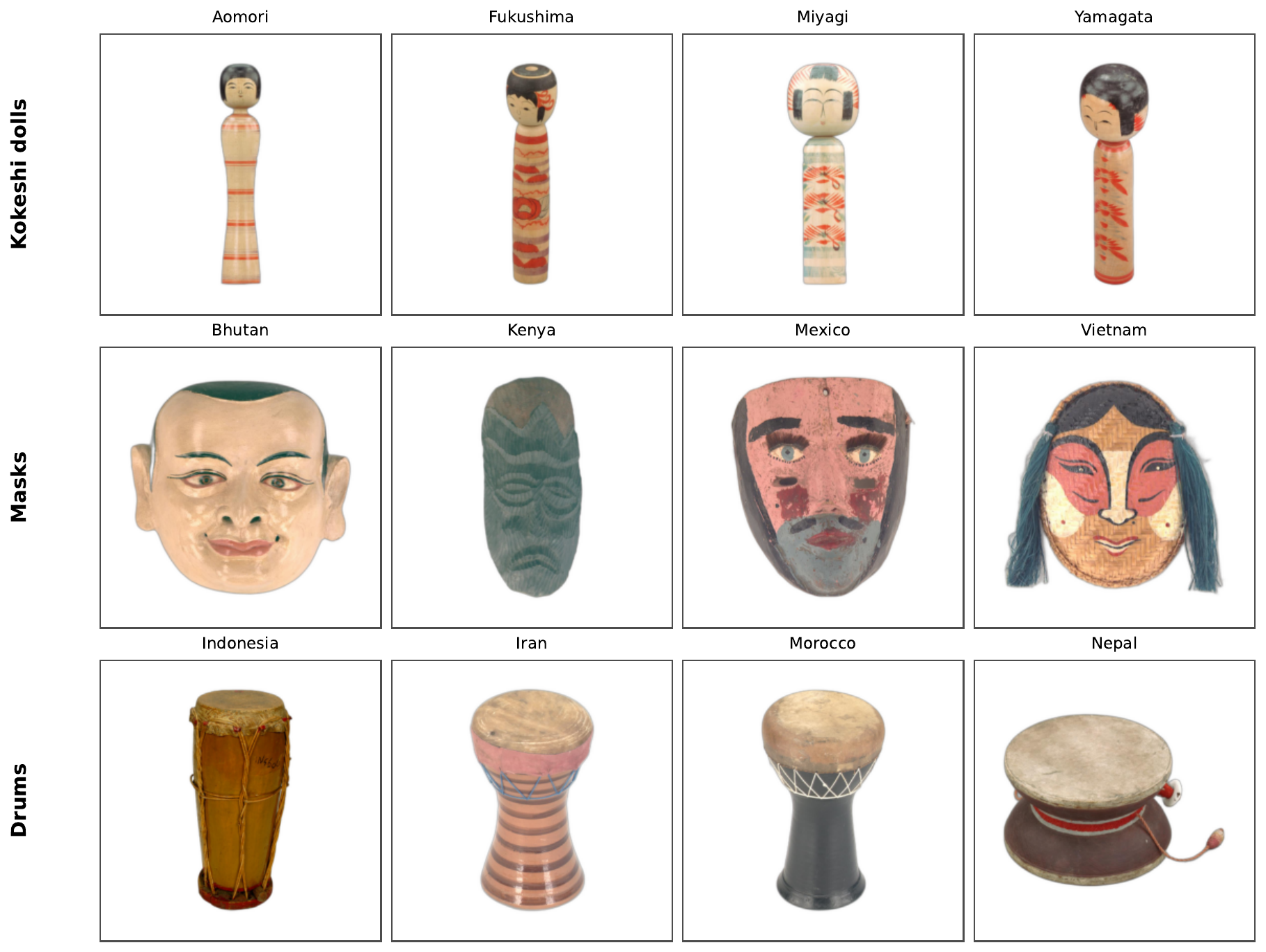}
    \caption{Examples of selected regions in each artifact category.
    For each category, four target regions were selected, 
    and five representative images from each region were used for evaluation.
    The examples show that artifacts in the same category share similar overall structures, 
    while local appearances such as patterns, ornaments, and facial details vary across regions.}
  \label{fig:dataset_examples}
\end{figure}

\section{Inpainting Based Reconstruction Analysis of Museum Artifact Features}
\label{sec:method}
\begin{figure}[tb]
  \centering
  \includegraphics[width=\textwidth]{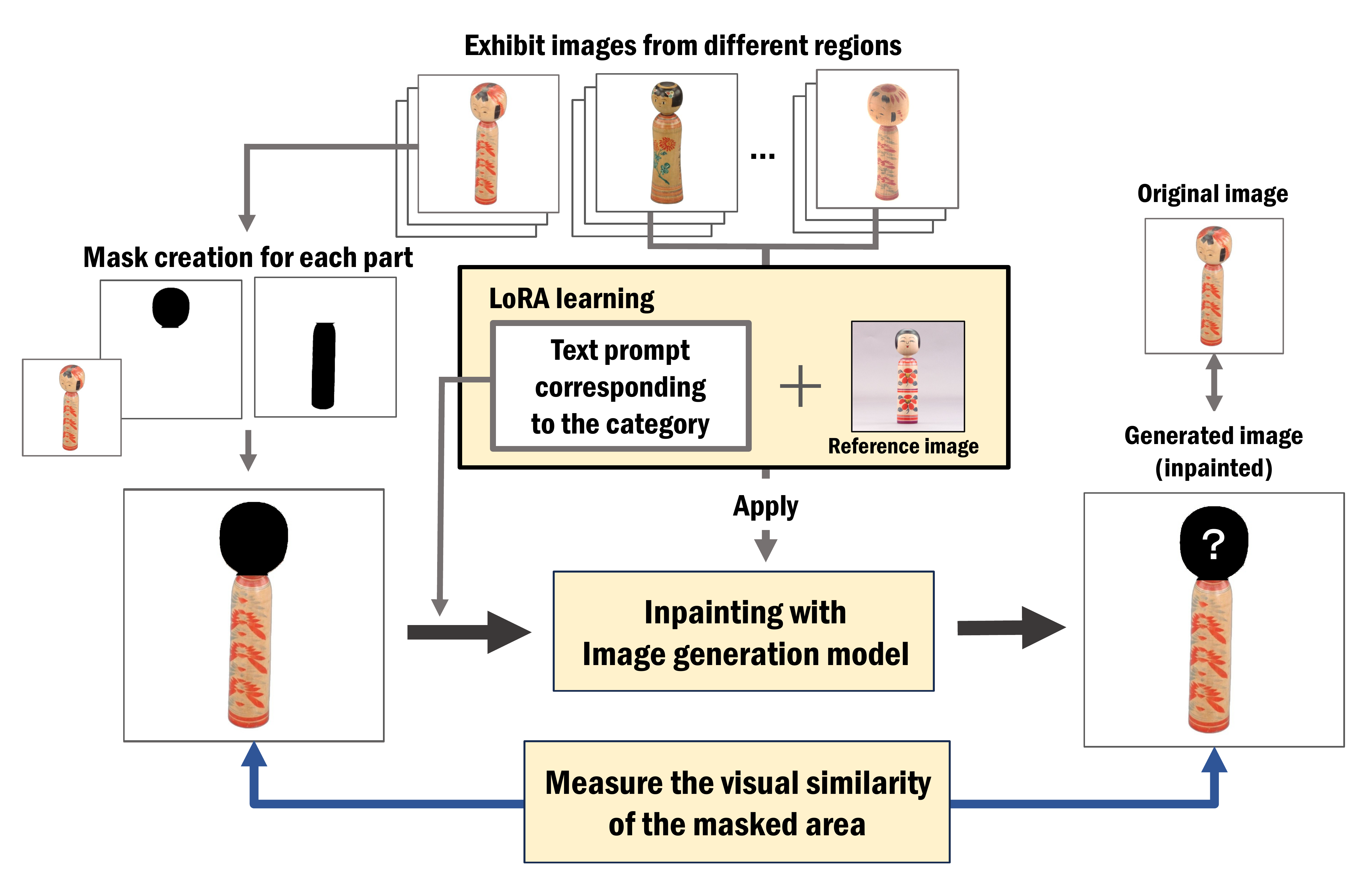}
  \caption{Overview of the proposed method.
        For each target region $r_i$, a LoRA model is trained using data from all other regions.
        The model is then used to inpaint masked parts of artifacts from the target region.
        The LPIPS distance between the inpainted image and the original image
        is measured as the local visual unexpectedness score.}
  \label{fig:pipeline}
\end{figure}

The proposed method consists of four steps.
First, we generate masks for parts of each artifact.
Second, we fine-tune an image generation model with LoRA using data that exclude a specific region.
Third, we use the trained model to inpaint the masked parts.
Fourth, we measure the difference from the original image using LPIPS
and calculate the unexpectedness score.
The whole pipeline is shown in Fig.~\ref{fig:pipeline}.

\subsection{Part Based Mask Design}
\label{mask}
In the proposed method, local parts used for inpainting are defined for each artifact category.
The target parts for each category are shown in Table~\ref{tab:target_parts}.
For each part, we manually set a rectangular region that fully covers the part,
which is used as the bounding box for that part.
In the actual inpainting process, we do not use the rectangle as it is.
Instead, we create a binary mask by removing unnecessary background areas
according to the outer silhouette of the artifact,
and use this as the input mask.

\begin{figure}[tb]
    \centering
    \includegraphics[width=\textwidth]{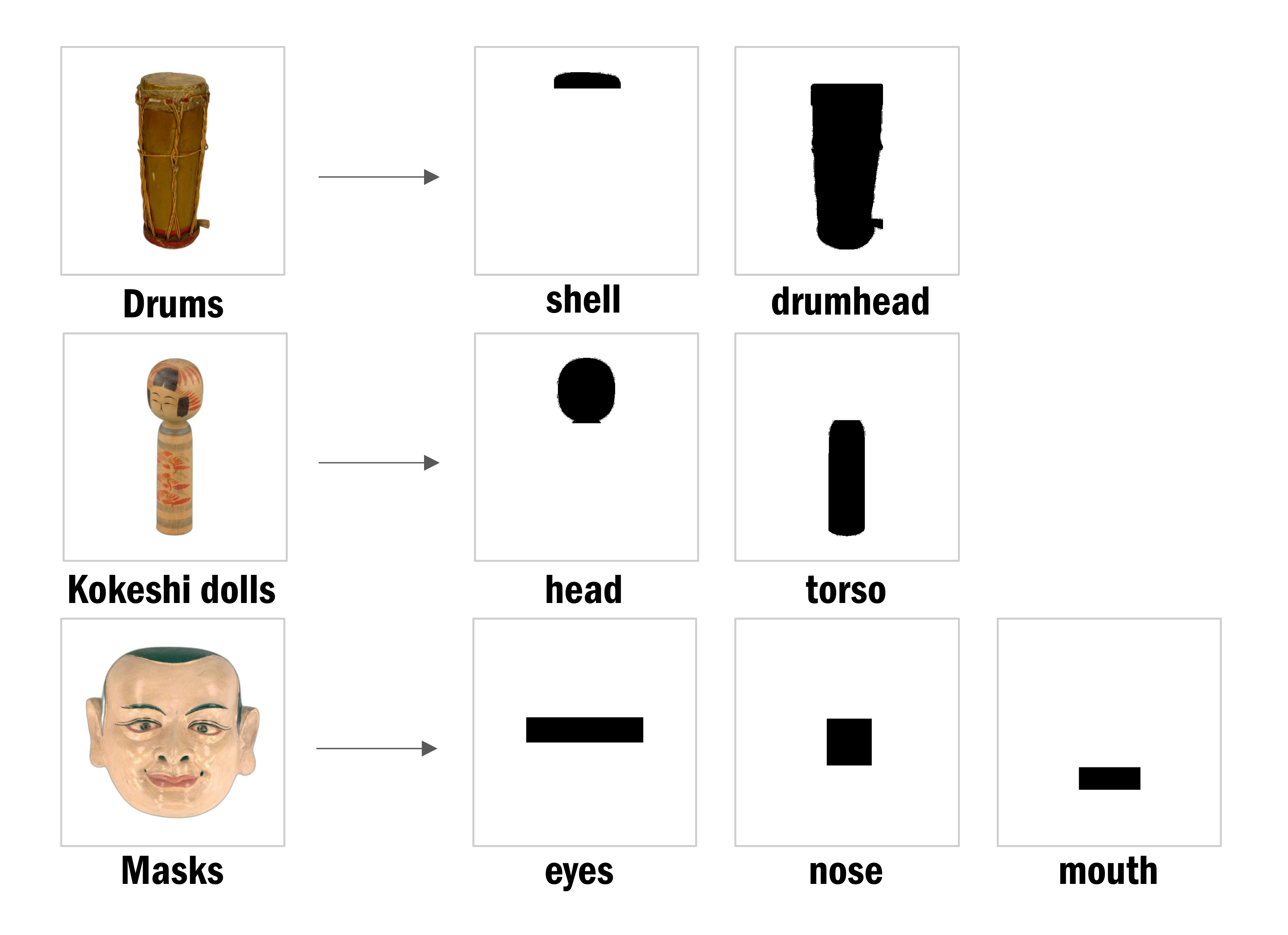}
    \caption{Examples of binary inpainting masks for each predefined artifact part.}
    \label{fig:mask_img}
\end{figure}

Fig.~\ref{fig:mask_img} shows examples of the binary inpainting masks used in this study.

\begin{table}[tb]
    \caption{Target parts defined for each artifact category.}
    \label{tab:target_parts}
    \centering
        \begin{tabular}{ll}
        \hline \hline
        Artifact category & Target part \\
        \hline
        Kokeshi doll & head, body \\
        Mask & eyes, nose, mouth \\
        Drum & face, body \\
        \hline
        \end{tabular}
\end{table}

\subsection{Estimating Visual Unexpectedness through LoRA Training and Inpainting}
In this study, we use Stable Diffusion~\cite{Rombach2022sd},
an image generation method based on diffusion models.
Stable Diffusion can generate high resolution images,
and it also has the advantage that it can be fine-tuned relatively easily
from an existing model~\cite{gal2023textual,Ruiz2023DreamBooth}.
To learn the general appearance of artifacts while excluding features from the target region,
we fine-tune the model using LoRA~\cite{hu2022lora}.
LoRA was introduced to reduce memory use and computation cost
when fine-tuning large language models.
In this study, we use LoRA training so that the model learns general appearance patterns
of Minpaku artifacts only from images outside the target region.
These learned patterns do not directly include region-specific features of the target region.

\subsubsection{Region-Excluded LoRA Fine-Tuning}

For each evaluation target region $r_i$, we build a LoRA model
by excluding images from that region from the training data.
Specifically, from all images in category $C$, we remove images from region $r_i$
and fine-tune Stable Diffusion using only the remaining images.
For example, when evaluating \textit{kokeshi dolls} from Yamagata Prefecture,
we exclude images from Yamagata and train LoRA using kokeshi images from other regions.
The model obtained in this way has not directly learned visual features specific to region $r_i$.
Therefore, in this study, we treat this model as a model that represents
visual expectations for the category formed without knowing the features of that region.
We use LoRA for training and use \texttt{Stable Diffusion v1.5} as the base model.
The image resolution is $512\times512$, the LoRA rank is 32,
and the number of training steps is 8,000.
For text prompts during training, we use simple expressions that describe each category.
The specific training prompts are shown in Table~\ref{tab:prompt}.
Because special handling was needed for \textit{kokeshi dolls}, it is explained in the next subsection.

Through this region excluded LoRA training,
we build a generative model for each region that represents the general appearance distribution
expected when that region is excluded.

\subsubsection{Special Handling for Kokeshi Dolls}

For the kokeshi doll category, we prepared the dataset and trained the model
in a different way from the other artifact categories.
For drums and masks, as described above,
we trained LoRA using images paired with simple text prompts.
However, when we trained kokeshi images with the prompt \textit{a kokeshi doll},
we found a problem: a face appeared in the generated result during inpainting
regardless of which part of the image was masked.
The kokeshi images used for training had similar camera angles,
and the face was placed near the center in many images.
Because of this, LoRA training seems to have strongly linked the text \textit{kokeshi doll}
with a structure in which a face exists near the center.
To reduce this problem, we split each kokeshi image at around 20--30\% from the top.
This produced a square head image from the upper part
and a square body image from the lower part.
We also assigned part specific prompts to these images, as shown in Table~\ref{tab:prompt},
and added them to the training data.
With this setup, the model learns \textit{kokeshi dolls} as structured objects
with a head and a body.
By using prompts that depend on the masked position during inpainting,
the problem of unnaturally generated faces was solved.
In this study, we apply this handling only to the kokeshi doll category
and conduct the following experiments.

\begin{table*}[tb]
    \centering
    \caption{Prompts used for training and inpainting.}
    \label{tab:prompt}
    \begin{tabular}{c|c|l}
        \hline \hline
        Purpose & Category (part) & Prompt \\ \hline

        \multirow{6}{*}{LoRA training} &
        kokeshi doll (head) &
        kokeshi doll's round head \\ &
        kokeshi doll (body) &
        kokeshi doll's cylindrical body \\  &
        kokeshi doll (overall) &
        a kokeshi doll with a round head \\ &&
        and a cylindrical body \\ &
        drum &
        a drum \\ &
        mask &
        a mask \\ \hline

        \multirow{3}{*}{Inpainting} &
        \multirow{3}{*}{All targets} &
        In addition to the above prompt \\ &&
        centered, isolated on a plain \\ &&
        white background, studio lighting \\ \hline

    \end{tabular}
\end{table*}

\subsection{Inpainting}
Given a target image from region $r_i$,
we use the model $M_{-r_i}$ to inpaint each part mask.
Specifically, we use ControlNet and input the masked image.
The generated content is made to match the surrounding artifact,
while being sampled from the distribution learned by $M_{-r_i}$.
For the text prompt used during inpainting, 
we employ the category- or part-specific prompt used during LoRA training and additionally append a short auxiliary prompt describing image quality and background conditions. 
The complete prompt configuration is shown in Table~\ref{tab:prompt}.
Because generation with a diffusion model can vary depending on the random seed,
we perform inpainting five times for each mask ($N=5$)
and average the LPIPS scores.

\subsection{LPIPS-Based Unexpectedness Score}
We measure the difficulty of inpainting, or the difference from the generated result,
using LPIPS~\cite{zhang2018LPIPS}.
For each inpainted image, we calculate LPIPS between the original image
and the generated image in the masked part.
\section{Results and Discussion}
\label{result}

In this section, we analyze the reconstruction difficulty scores obtained by the proposed method and the corresponding inpainting results. 
The goal is to examine what kinds of local visual features are difficult for the region-excluded models to reconstruct. 
In this study, the LPIPS-based difference is not interpreted as a direct measure of human subjective unexpectedness, 
but as reconstruction difficulty with respect to the common appearance distribution learned by the generative model. 
Therefore, we discuss both the overall tendencies of the scores and qualitative examples of the generated results.

\subsection{Overall Reconstruction Difficulty}

We first examine the overall tendency of reconstruction difficulty across artifact categories and target parts. 
Table~\ref{tab:lpips_part} shows the mean and standard deviation of LPIPS scores for each category and part. 
A higher score indicates a larger perceptual difference between the original part and the inpainted result, 
suggesting that the appearance of the part was more difficult for the region-excluded model to reconstruct.

\begin{table}[tb]
    \caption{Reconstruction difficulty scores for each category and target part.}
    \label{tab:lpips_part}
    \centering
    \begin{tabular}{llrr}
    \hline \hline
    \hline
    category & part name & mean lpips & std lpips\\
    \hline
        Drum & shell & 0.642 & 0.058\\
        Drum & drumhead & 0.488 & 0.085\\
        Kokeshi & torso & 0.544 & 0.046\\
        Kokeshi & head & 0.559 & 0.044\\
        Mask & mouth & 0.563 & 0.054\\
        Mask & eyes & 0.543 & 0.065\\
        Mask & nose & 0.546 & 0.079\\
    \hline
    \end{tabular}
\end{table}

Table~\ref{tab:difficulty_scores} shows the mean LPIPS score for each artifact category and target region. 
This table indicates which regional groups were more difficult for the region excluded models to reconstruct. 
The values are averaged over the evaluation images and target parts for each region.

    \begin{table}[tb]
    \caption{Mean reconstruction difficulty scores for each category and target region.}
    \label{tab:difficulty_scores}
    \centering
    \begin{tabular}{llr}
    \hline \hline
        \hline
        Category & Region & Mean LPIPS \\
        \hline
        Drum & Morocco & 0.597 \\
        Drum & Indonesia & 0.537 \\
        Drum & Nepal & 0.568 \\
        Drum & Iran & 0.550 \\
        Mask & Mexico & 0.571 \\
        Mask & Vietnam & 0.617 \\
        Mask & Bhutan & 0.526 \\
        Mask & Kenya & 0.501 \\
        Kokeshi & Miyagi & 0.570 \\
        Kokeshi & Fukushima & 0.552 \\
        Kokeshi & Aomori & 0.548 \\
        Kokeshi & Yamagata & 0.529 \\
    \hline
    \end{tabular}
\end{table}

As shown in Table~\ref{tab:lpips_part}, drum shell regions showed the highest mean LPIPS score among the target parts. 
This suggests that drum shell regions, which contain colors, cords, and ornaments, were more difficult to reconstruct than simpler parts such as drumheads or local facial components of \textit{masks}. 
However, this tendency should be interpreted carefully. 
The shell region of a drum is larger than many other target parts, 
and larger masks provide less surrounding context for inpainting. 
Therefore, the high score for drum shells may reflect not only region-specific visual features, 
but also the inherent difficulty caused by the size of the masked region. 

In the region-wise results shown in Table~\ref{tab:difficulty_scores}, 
Vietnamese \textit{masks} and Moroccan drums showed relatively high mean scores, 
whereas Kenyan \textit{masks} and Yamagata \textit{kokeshi dolls} showed relatively low mean scores. 
These tendencies suggest that the proposed score reflects not only artifact categories, 
but also differences in local decorations, colors, and structures across regions.

\subsection{Region-Specific Inpainting Results}

We next examine concrete inpainting results, focusing on artifacts with relatively high reconstruction difficulty scores. 
Table~\ref{tab:high_low_inpainting_results} shows representative high- and low-scoring examples
for each artifact category.
For each example, we present the original image, the mask, and the inpainted result. 
In this analysis, we focus not only on the magnitude of the LPIPS score, but also on what local features were preserved and 
what features were replaced with more common appearances in the generated results.

\subsection{Inpainting Results for Region-Specific Visual Features}

In the inpainting results, the generated drums are visually plausible as drums however, 
the body color changes to reddish-brown tones, and the string structure is replaced with vertical strings or different decorative arrangements. 
This indicates that the region-excluded model can generate a plausible drum-like appearance, 
However, the state of the strings and patterns seen in the original image was perfectly reproduced.

This result also illustrates that high reconstruction difficulty does not necessarily correspond to a visually broken generation. 
Rather, the model completes the masked part in a visually natural way, 
but replaces the original region-specific appearance with a more common appearance based on its learned distribution.

It should also be noted that the body region of drums occupies a relatively large area compared with other target parts. 
Since inpainting relies on the surrounding context, a large masked region provides less local context for reconstruction. 
Therefore, the high LPIPS scores for drum bodies may reflect not only region-specific color and ornament differences, 
but also the inherent difficulty caused by the size of the masked region. 
This needs to be considered when interpreting the score.

In the mask category, artifacts from Vietnam showed the highest regional mean LPIPS score. 
However, for facial parts such as the eyes, nose, and mouth, 
the generated results often fit the surrounding facial context and produced visually plausible local components. 
For example, in the nose part of Vietnam, 
the generated eyes do not perfectly match the original expression or decoration, but they are still consistent with the overall face.
This suggests that even for masks with distinctive decorations, 
the surrounding facial structure provides strong contextual cues, allowing the model to generate locally plausible parts.

In the kokeshi category, variations in the generated results were observed for both the head and body parts. 
For the head region, stylized representations of black hair were often reproduced successfully, even when the original image depicted the hair as a simple black shape. 
For the body region, when the original image contained specific motifs, such as flowers with branches or plum-blossom-like decorations, these motifs were rarely reproduced directly. 
Instead, they were often replaced with more common floral patterns or stripe-like designs. 
Conversely, when the original body contained only a small amount of decoration, 
the generated result sometimes introduced additional ornaments or stripe patterns that were not present in the original image.

These results show that the region-excluded models can generate general shapes and local parts of each artifact category, 
while specific decorations, colors, string arrangements, and stylized representations in the original images tend to be replaced with more common appearances. 
Such replacements provide cues that the local features of the target part were difficult to reconstruct from the appearance distribution learned by the model.

\subsection{High- and Low-Scoring Examples}

To further examine how reconstruction difficulty scores correspond to actual generation results, 
we compare high-scoring and low-scoring examples. 
Tab~\ref{tab:high_low_inpainting_results} shows examples with high and low LPIPS scores. 

\begin{table}[tb]
        \centering 
        \caption{Inpainting results for high- and low-scoring examples in each artifact category.} 
        \label{tab:high_low_inpainting_results} 
        \small     
        \begin{tabular}{c|c|c|c|c|c}
        \hline\hline
        Category & Region & Original & Mask & Inpainted & LPIPS \\
        \hline

        Kokeshi dolls-torso
        & Yamagata
        & \begin{minipage}{0.17\columnwidth}
            \centering
            \scalebox{0.8}{\includegraphics[width=\linewidth]{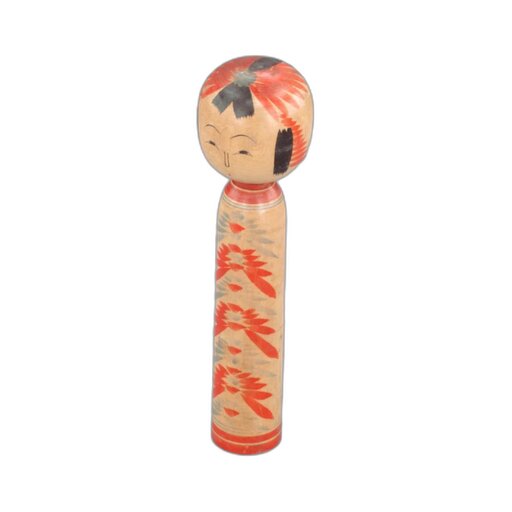}}
        \end{minipage}
        & \begin{minipage}{0.17\columnwidth}
            \centering
            \scalebox{0.8}{\includegraphics[width=\linewidth]{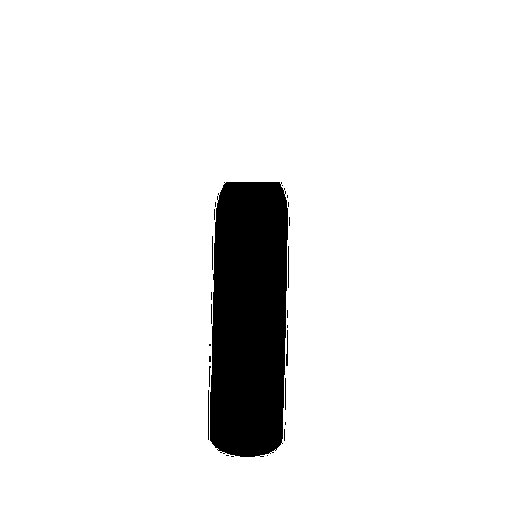}}
        \end{minipage}
        & \begin{minipage}{0.17\columnwidth}
            \centering
            \scalebox{0.8}{\includegraphics[width=\linewidth]{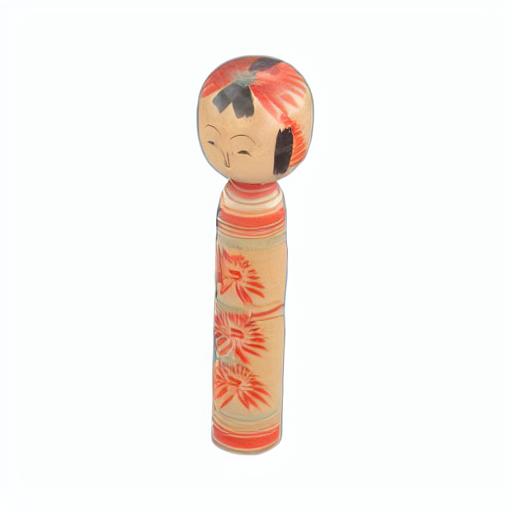}}
        \end{minipage}
        & 0.430 \\
        
        Kokeshi dolls-torso
        & Miyagi
        & \begin{minipage}{0.17\columnwidth}
            \centering
            \scalebox{0.8}{\includegraphics[width=\linewidth]{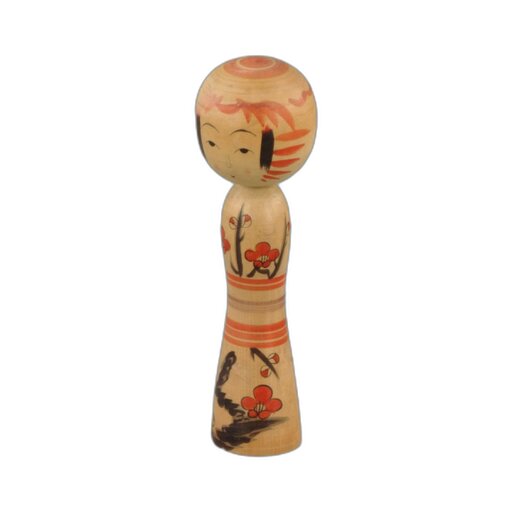}}
        \end{minipage}
        & \begin{minipage}{0.17\columnwidth}
            \centering
            \scalebox{0.8}{\includegraphics[width=\linewidth]{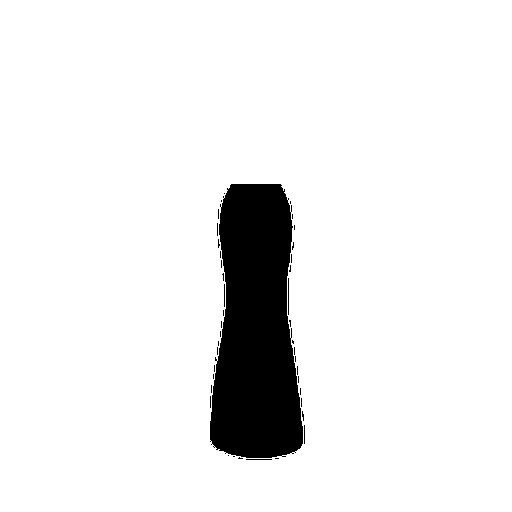}}
        \end{minipage}
        & \begin{minipage}{0.17\columnwidth}
            \centering
            \scalebox{0.8}{\includegraphics[width=\linewidth]{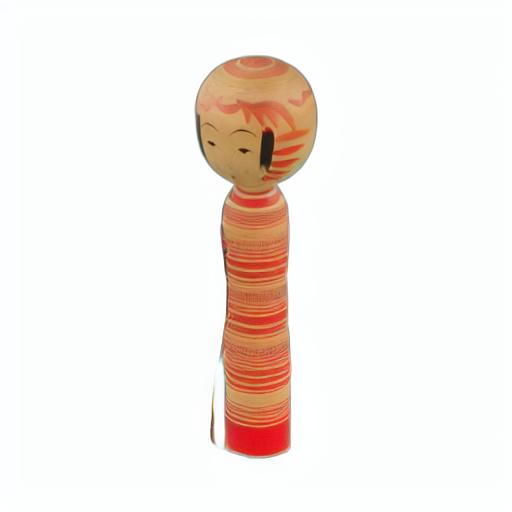}}
        \end{minipage}
        & 0.613 \\
        
        \hline

        Drums-drumhead
        & Iran
        & \begin{minipage}{0.17\columnwidth}
            \centering
            \scalebox{0.8}{\includegraphics[width=\linewidth]{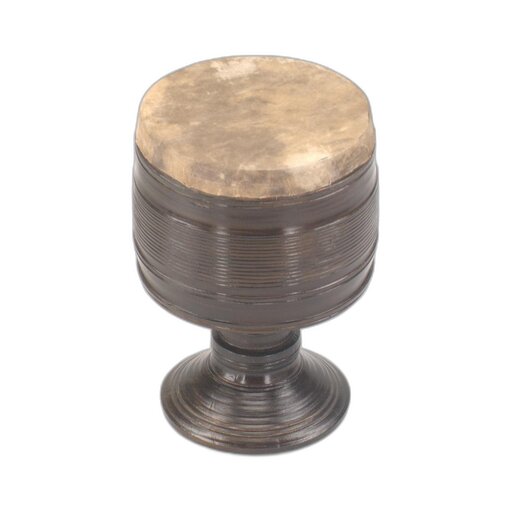}}
        \end{minipage}
        & \begin{minipage}{0.17\columnwidth}
            \centering
            \scalebox{0.8}{\includegraphics[width=\linewidth]{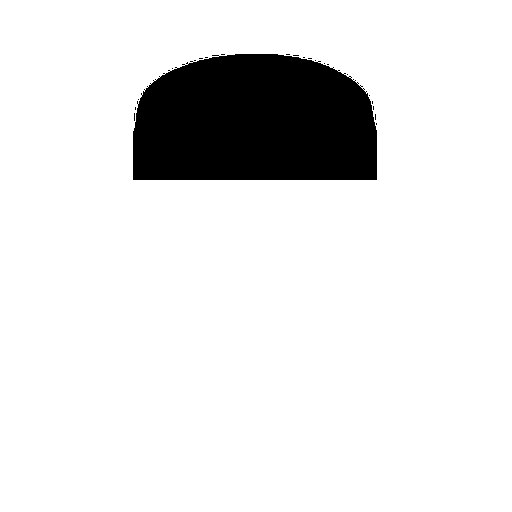}}
        \end{minipage}
        & \begin{minipage}{0.17\columnwidth}
            \centering
            \scalebox{0.8}{\includegraphics[width=\linewidth]{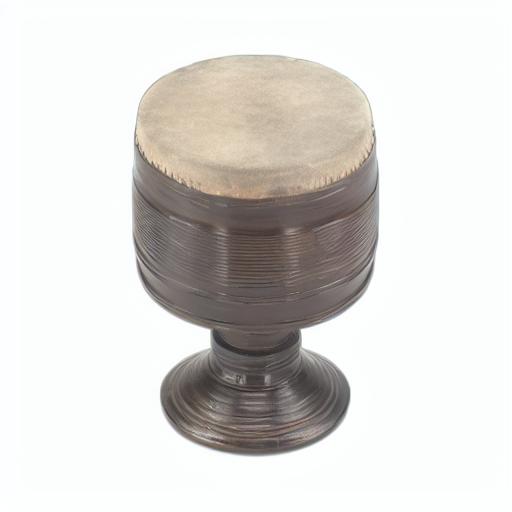}}
        \end{minipage}
        & 0.426 \\
        
        Drums-shell 
        & Indonesia
        & \begin{minipage}{0.17\columnwidth}
            \centering
            \scalebox{0.8}{\includegraphics[width=\linewidth]{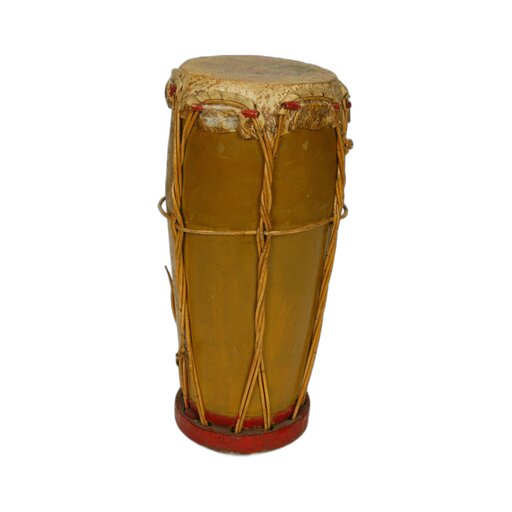}}
        \end{minipage}
        & \begin{minipage}{0.17\columnwidth}
            \centering
            \scalebox{0.8}{\includegraphics[width=\linewidth]{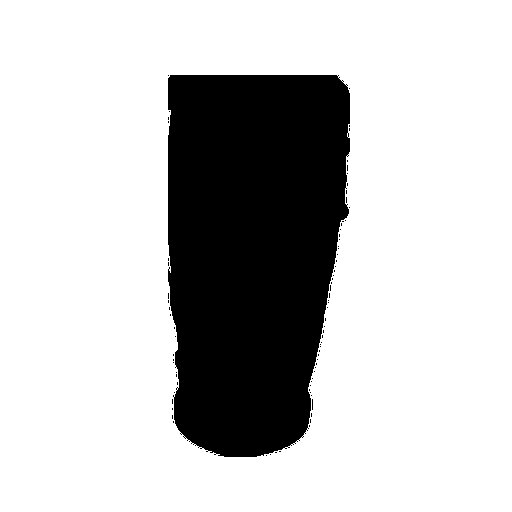}}
        \end{minipage}
        & \begin{minipage}{0.17\columnwidth}
            \centering
            \scalebox{0.8}{\includegraphics[width=\linewidth]{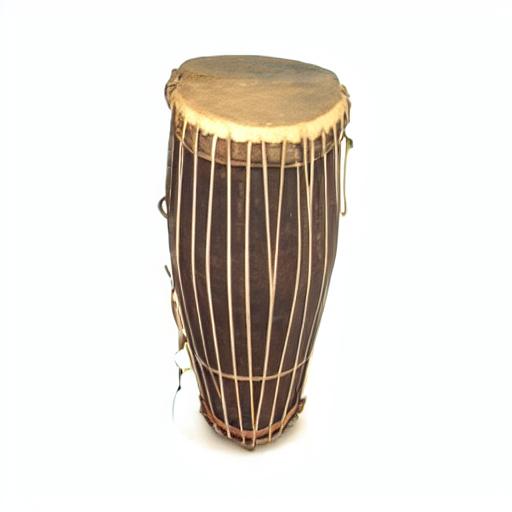}}
        \end{minipage}
        & 0.739 \\
        
        \hline
        Masks-nose
        & Kenya 
        & \begin{minipage}{0.17\columnwidth}
            \centering
            \scalebox{0.8}{\includegraphics[width=\linewidth]{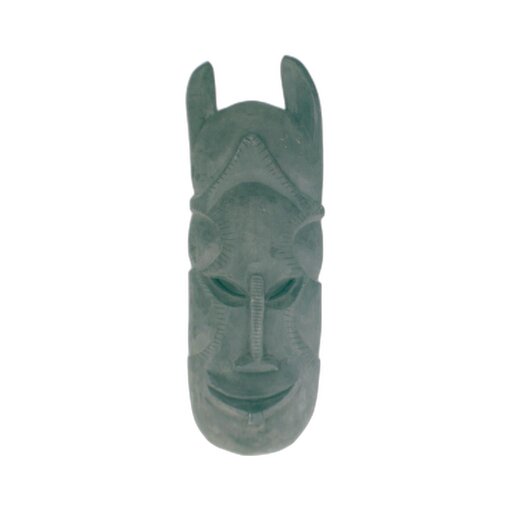}}
        \end{minipage}
        & \begin{minipage}{0.17\columnwidth}
            \centering
            \scalebox{0.8}{\includegraphics[width=\linewidth]{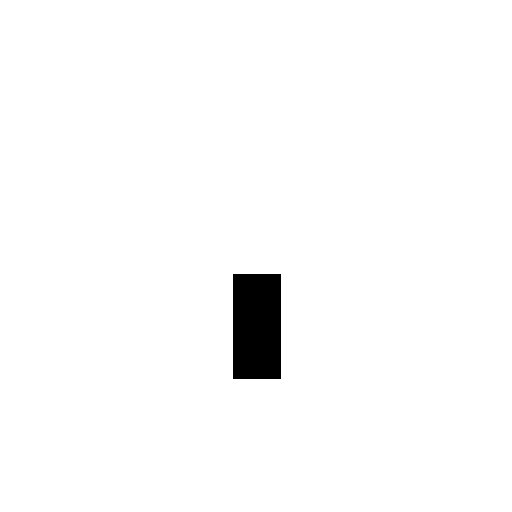}}
        \end{minipage}
        & \begin{minipage}{0.17\columnwidth}
            \centering
            \scalebox{0.8}{\includegraphics[width=\linewidth]{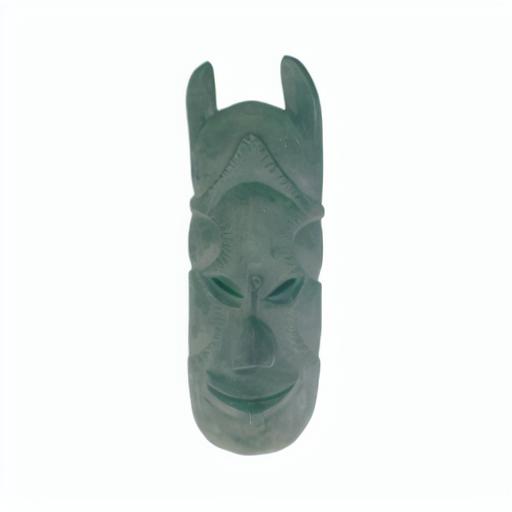}}
        \end{minipage}
        & 0.452 \\
        
        Masks-nose 
        & Vietnam
        & \begin{minipage}{0.17\columnwidth}
            \centering
            \scalebox{0.8}{\includegraphics[width=\linewidth]{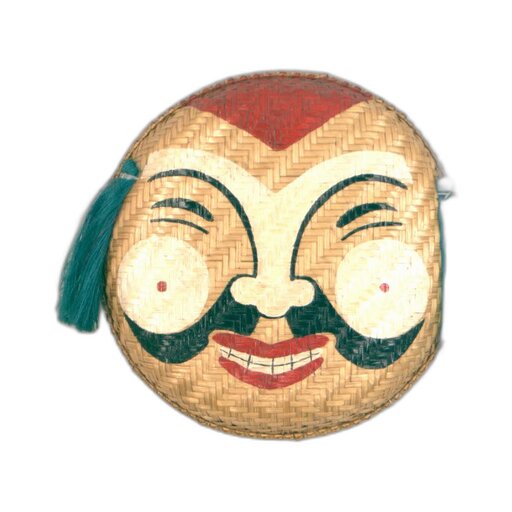}}
        \end{minipage}
        & \begin{minipage}{0.17\columnwidth}
            \centering
            \scalebox{0.8}{\includegraphics[width=\linewidth]{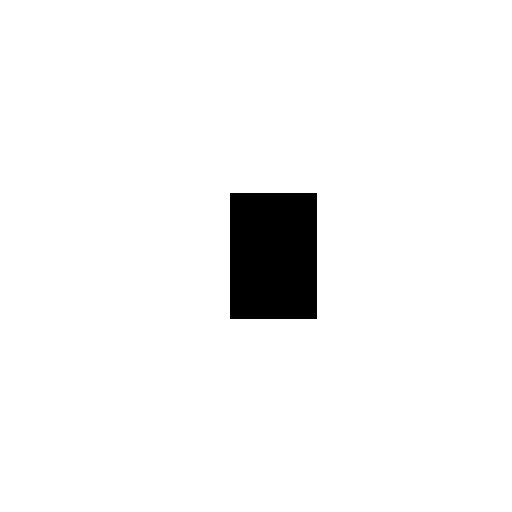}}
        \end{minipage}
        & \begin{minipage}{0.17\columnwidth}
            \centering
            \scalebox{0.8}{\includegraphics[width=\linewidth]{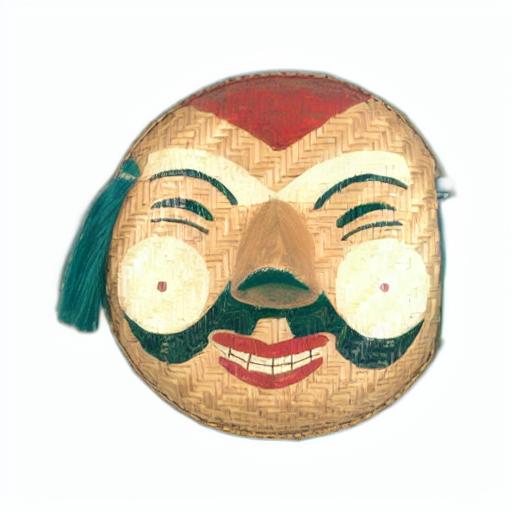}}
        \end{minipage}
        & 0.614 \\
        
        \hline
    \end{tabular}
\end{table}

To further examine how reconstruction difficulty scores correspond to actual generation results,
we compare high- and low-scoring examples.
Fig.~\ref{tab:high_low_inpainting_results} shows representative high- and low-scoring examples
for each artifact category.

As an example of a high-scoring case, a \textit{kokeshi doll} from Miyagi showed a large discrepancy between the original and generated body decorations. 
While the overall shape and color distribution were preserved, 
the decorative pattern itself was transformed into a visually different arrangement during inpainting. 
This perceptual change in the body region led to a high LPIPS score.

In the high scoring drum example from Indonesia,
the overall drum structure was preserved,
but the number and arrangement of the tension cords around the drumhead changed.
While the original image has a structure in which the drumhead is supported by relatively few cords,
the inpainted result generated a larger number of cords.
Thus, the generated image remained visually plausible as a drum,
but the local structural details differed from the original image,
resulting in a high LPIPS score.

In the high scoring \textit{mask} example from Vietnam,
the face has distinctive painted patterns.
When the nose region was masked and inpainted,
the model generated a locally plausible nose like shape.
However, the original painted pattern and decoration were not preserved.
As a result, although the generated part was not visually broken,
it differed from the original local appearance and produced a high LPIPS score.

Next, we examine the low scoring examples.
In the \textit{kokeshi doll} example from Yamagata,
the body contains stripe like and repeated decorative patterns
that are commonly observed in \textit{kokeshi dolls}.
The inpainted result also generated a similar type of body decoration,
and the difference from the original pattern was relatively small.

In the drum example from Iran,
the drumhead region was inpainted.
Although the generated drumhead showed a slight color difference from the original,
it was visually consistent with the surrounding structure
and appeared natural as a drumhead.
Therefore, the difference from the original image was small,
leading to a low LPIPS score.

In the \textit{mask} example from Kenya,
the nose region was inpainted.
This artifact has a relatively uniform color appearance,
and the nose region does not contain complex decorations or highly distinctive local patterns.
The inpainted result similarly generated a simple nose region
that was consistent with the surrounding face,
which resulted in a low LPIPS score.

These comparisons show that high LPIPS scores do not simply indicate failed generation.
Rather, in high scoring examples, local patterns, decorations, and structures in the original image
are replaced with more common appearances learned by the region excluded model.
In contrast, low scoring examples are completed within the learned appearance distribution,
and the generated parts remain close to the original local appearance.
Therefore, the LPIPS based reconstruction difficulty score can be used as a cue
for examining how strongly local visual features deviate from the general appearance distribution
learned by the model, rather than as a direct measure of generation quality.

\subsection{Limitations}

The proposed score should be interpreted as a measure of reconstruction difficulty 
for a region-excluded generative model rather than a direct measure of human subjective unexpectedness. 
While perceptual differences captured by LPIPS may correlate with visually distinctive regional characteristics, 
further user studies would be required to establish a direct relationship with human perception.

In addition, because the score is derived from an inpainting process, 
it is influenced by the amount of contextual information available during reconstruction. 
Consequently, larger masked regions may yield higher scores than smaller regions, 
independent of the strength of regional characteristics.
Finally, the present evaluation was conducted using available datasets 
whose sizes vary across categories and regions. 
Future work will investigate the robustness of the proposed score using larger and more balanced datasets 
and will further examine its relationship to human judgments of regional distinctiveness.

\section{Conclusion}
This study proposed an inpainting-based framework for analyzing local visual characteristics of museum artifacts through reconstruction difficulty. 
For each target region, a region-excluded LoRA model was trained using artifact images from other regions, and the model was used to inpaint masked regions of artifacts from the excluded region. The perceptual difference between the original and reconstructed regions was then quantified using LPIPS.
Experiments on \textit{kokeshi dolls}, \textit{drums}, and \textit{masks} demonstrated that the proposed method can identify local features that are difficult to reproduce from appearance distributions learned from other regions. While common shapes and decorative elements were generally reconstructed successfully, distinctive regional characteristics were often replaced by more typical appearances. These observations suggest that reconstruction difficulty captures aspects of regional visual distinctiveness that are not easily explained by shared appearance patterns.
The proposed framework provides a data-driven approach for exploring regional characteristics in museum artifact collections without requiring manual annotation of visual features. Future work includes investigating the relationship between reconstruction difficulty and human perception, improving mask design and score normalization, and extending the method to a wider range of cultural heritage materials.

\begin{credits}
\subsubsection{\ackname} This work was supported by JSPS KAKENHI Grant Numbers JP25K03229, JP26K23922, JP25K03228, and JP24K03228.

\subsubsection{\discintname}
The authors have no competing interests to declare that are relevant to the content of this article.

\subsubsection{Generative AI Usage Disclosure.}
During the preparation of this work, the authors used Gemini 3.5 Flash in order to improve the English phrasing and readability of the manuscript. After using this tool, the authors reviewed and edited the content as needed and take full responsibility for the content of the publication.
\end{credits}
\bibliographystyle{splncs04}
\bibliography{references}

\end{document}